\documentclass{article}

\usepackage{microtype}
\usepackage{graphicx}
\usepackage{subfigure}
\usepackage{booktabs} %

\usepackage[pdfa]{hyperref}
\hypersetup{
    pdftitle = {Differentiable Top-k Classification Learning},
    pdflang = {en-US},
}
\usepackage{url}

\usepackage{wrapfig}

\usepackage{amsmath}
\usepackage{xcolor}

\usepackage{amsmath,amsfonts,bm}

\def\eqref#1{equation~\ref{#1}}

\def\1{\bm{1}}

\def\mP{{\bm{P}}}

\DeclareMathAlphabet{\mathsfit}{\encodingdefault}{\sfdefault}{m}{sl}
\SetMathAlphabet{\mathsfit}{bold}{\encodingdefault}{\sfdefault}{bx}{n}

\newcommand{\softmax}{\mathrm{softmax}}

\DeclareMathOperator*{\argmax}{arg\,max}

\usepackage{pgf}

\usepackage{amssymb}%
\usepackage{pifont}%
\newcommand{\cmark}{\ding{51}}%
\newcommand{\xmark}{\ding{55}}%

\newcommand{\revA}[1]{{#1}}

\newcommand{\revC}[1]{{#1}}
\newcommand{\revD}[1]{{#1}}

\usepackage[accepted]{icml2022}

\usepackage{printlen}

\usepackage{amsthm}

\icmltitlerunning{Differentiable Top-$k$ Classification Learning}

\begin{document}

\twocolumn[
\icmltitle{Differentiable Top-$k$ Classification Learning}

\begin{icmlauthorlist}
\icmlauthor{Felix Petersen}{kn}
\icmlauthor{Hilde Kuehne}{f,m}
\icmlauthor{Christian Borgelt}{s}
\icmlauthor{Oliver Deussen}{kn}
\end{icmlauthorlist}

\icmlaffiliation{kn}{University of Konstanz}
\icmlaffiliation{s}{University of Salzburg}
\icmlaffiliation{f}{University of Frankfurt}
\icmlaffiliation{m}{MIT-IBM Watson AI Lab}

\icmlcorrespondingauthor{Felix Petersen}{felix.petersen@uni.kn}

\icmlkeywords{Machine Learning, ICML}

\vskip 0.3in
]

\printAffiliationsAndNotice{}  %

\begin{abstract}

The top-$k$ classification accuracy is one of the core metrics in machine learning.
Here, $k$ is conventionally a positive integer, such as $1$ or $5$, leading to top-$1$ or top-$5$ training objectives.
In this work, we relax this assumption and optimize the model for multiple $k$ simultaneously instead of using a single $k$.
Leveraging recent advances in differentiable sorting and ranking, we propose a differentiable top-$k$ cross-entropy classification loss.
This allows training the network while not only considering the top-$1$ prediction, but also, e.g., the top-$2$ and top-$5$ predictions.  
We evaluate the proposed loss function for fine-tuning on state-of-the-art architectures, as well as for training from scratch.
We find that relaxing $k$ does not only produce better top-$5$ accuracies, but also leads to top-$1$ accuracy improvements.
When fine-tuning publicly available ImageNet models, we achieve a new state-of-the-art for these models.

\end{abstract}

\section{Introduction}

Classification is one of the core disciplines in machine learning and computer vision.
The advent of classification problems with hundreds or even thousands of classes let the top-$k$ classification accuracy establish as an important metric, i.e., one of the top-$k$ classes has to be the correct class.
Usually, models are trained to optimize the top-$1$ accuracy; and top-$5$ etc.~are used for evaluation only.
Some works \citep{lapin2016loss,berrada2018smooth} have challenged this idea and proposed top-$k$ losses, such as a smooth top-$5$ margin loss.
These methods have demonstrated superior robustness over the established top-$1$ softmax cross-entropy in the presence of additional label noise \citep{berrada2018smooth}.
In standard classification settings, however, these methods have so far not shown improvements over the established top-$1$ softmax cross-entropy.

In this work, instead of selecting a single top-$k$ metric such as top-$1$ or top-$5$ for defining the loss,
we propose to specify $k$ to be drawn from a distribution $P_K$, which may or may not depend on the confidence of specific data points or on the class label.
Examples for distributions $P_K$ are $[.5, 0, 0, 0, .5]$ ($50\%$ top-$1$ and $50\%$ top-$5$), $[.1, 0, 0, 0, .9]$ ($10\%$ top-$1$ and $90\%$ top-$5$), and $[.2, .2, .2, .2, .2]$ ($20\%$ top-$k$ for each $k$ from $1$ to $5$).
\revC{Note that, when $k$ is drawn from a distribution, this is done sampling-free as we can compute the expectation value in closed form.} %

Conventionally, given scores returned by a neural network, softmax produces a probability distribution over the top-$1$ rank.
Recent advances in differentiable sorting and ranking \citep{Grover2019-NeuralSort, prillo2020softsort, Cuturi2019-SortingOT, Petersen2021-diffsort} provide methods for generalizing this to  probability distributions over all ranks represented by a matrix $\mP$.
Based on differentiable ranking, multiple differentiable top-$k$ operators have recently been proposed.
They found applications in differentiable $k$-nearest neighbor algorithms, differentiable beam search, attention mechanisms, and differentiable image patch selection \citep{cordonnier2021differentiable}.
In these areas, integrating differentiable top-$k$ improved results considerably by creating a more natural end-to-end learning setting.
However, to date, none of the differentiable top-$k$ operators have been employed as neural network losses for top-$k$ classification learning with $k>1$.

Building on differentiable sorting and ranking methods, we propose a new family of differentiable top-$k$ classification losses where $k$ is drawn from a probability distribution.
We find that our top-$k$ losses improve not only top-$k$ accuracies, but 
also top-$1$ accuracy on multiple learning tasks.

We empirically evaluate our method using four differentiable sorting and ranking methods on the CIFAR-100 \citep{Krizhevsky2009_cifar10}, the ImageNet-1K \citep{deng2009imagenet}, and the ImageNet-21K-P \citep{ridnik2021imagenet} data sets.
Using CIFAR-100, we demonstrate the capabilities of our losses to train models from scratch.
On ImageNet-1K, we demonstrate that our losses are capable of fine-tuning existing models and achieve a new state-of-the-art for publicly available models on both top-$1$ and top-$5$ accuracy.
We benchmark our method on multiple recent models and demonstrate that our proposed method consistently outperforms the baselines for the best two differentiable sorting and ranking methods.
With ImageNet-21K-P, where many classes overlap (but only one is the ground truth), we demonstrate that our losses are scalable to more than $10\,000$ classes and achieve improvements of over $1\%$ with only last layer fine-tuning.

Overall, while the performance improvements on fine-tuning are rather limited (because we retrain only the classification head), they are consistent and can be achieved without the large costs of training from scratch.
The absolute $0.2\%$ improvement that we achieve on the ResNeXt-101 32x48d WSL top-$5$ accuracy corresponds to an error reduction by approximately $10\%$, and can be achieved at much less than 
the computational cost of (re-)training the full model in the first place.

We summarize our contributions as follows:
\vspace*{-.45em}
\begin{itemize}
\setlength{\itemindent}{-1.em}
\itemsep0pt
    \item We derive a novel family of top-$k$ cross-entropy losses and relax the assumption of a fixed $k$.
    \item We find that they improve both top-$k$ and top-$1$ accuracy.
    \item We demonstrate that our losses are scalable to more than $10\,000$ classes.
    \item We propose splitter selection nets, which require fewer layers than existing selection nets.
    \item We achieve new state-of-the-art results (for publicly available models) on ImageNet1K.
\end{itemize}
\vspace*{-.5em}

\begin{figure*}[h]
    \centering
    \includegraphics[width=.975\linewidth]{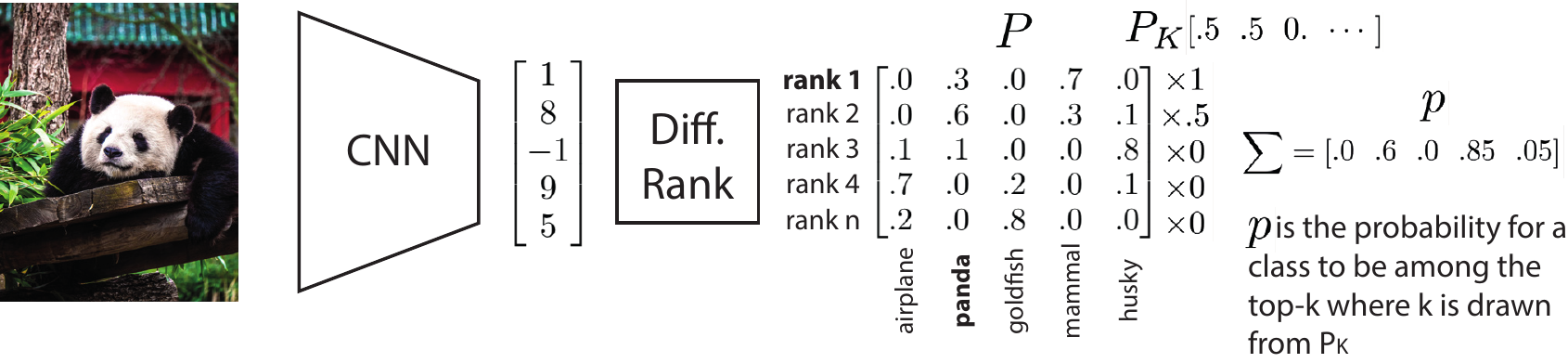}
    \caption{
        Overview of the proposed architecture:
        A CNN predicts scores for an image, which are then ranked by a differentiable ranking algorithm returning the probability distribution for each rank in matrix $\mP$.
        The rows of this distribution correspond to ranks, and the columns correspond to the respective classes.
        In the example, we use a $50\%$ top-$1$ and $50\%$ top-$2$ loss%
        \revA{%
        , i.e., $P_K=[.5, .5, 0, 0, 0]$. Here, the $k$th value refers to the top-$k$ component, which is satisfied if the prediction is at \textit{any} of \hbox{rank-$1$} to rank-$k$.
        Thus, the weights for the different ranks can be computed via a cumulative sum and are $[1, .5, 0, 0, 0]$.%
        }
        The correspondingly weighted sum of rows of $\mP$ yields the probability distribution~$p$, which can then be used in a cross-entropy loss.  
        Photo by Chris Curry on Unsplash.
    }
    \label{fig:panda-overview}
\end{figure*}

\section[Background: Differentiable Sorting and Ranking]{Background: Differentiable Sorting and Ranking}

\label{sec:background-diff-sort-and-rank}

We briefly review NeuralSort, SoftSort, Optimal Transport Sort, and Differentiable Sorting Networks.
We omit the fast differentiable sorting and ranking method~\citep{Blondel2020-FastSorting} and the relaxed Bubble sort algorithm~\cite{petersen2021learning} as they do not provide relaxed permutation matrices / probability scores, but rather only sorted / ranked vectors.

\subsection{NeuralSort \& SoftSort}
\vspace*{-.25em}
To make the sorting operation differentiable, \citet{Grover2019-NeuralSort} proposed relaxing permutation matrices to unimodal row-stochastic matrices. 
For this, they use the softmax of pairwise differences of (cumulative) sums of the top elements.
They prove that this, for the temperature parameter approaching $0$, is the correct permutation matrix, and propose a variety of deep learning differentiable sorting benchmark tasks.
They propose a deterministic softmax-based variant, as well as a Gumbel-Softmax variant of their algorithm. 
\revD{Note that NeuralSort is not based on sorting networks.}

\citet{prillo2020softsort} build on this idea but simplify the formulation and provide SoftSort, a faster alternative to NeuralSort.
They show that it is sufficient to build on pairwise differences of elements of the vectors to be sorted instead of the cumulative sums.
They find that SoftSort performs approximately equivalent in their experiments to NeuralSort.

\subsection{Optimal Transport / Sinkhorn Sort}
\citet{Cuturi2019-SortingOT} propose an entropy regularized optimal transport formulation of the sorting operation.
They solve this by applying the Sinkhorn algorithm~\citep{Cuturi13Sinkhorn} and produce gradients via automatic differentiation rather than the implicit function theorem, which resolves the need of solving a linear equation system.
As the Sinkhorn algorithm produces a relaxed permutation matrix, we can also apply Sinkhorn sort to top-$k$ classification learning.

\subsection{Differentiable Sorting Networks}
\citet{Petersen2021-diffsort} propose differentiable sorting networks, a continuous relaxation of sorting networks.
\revA{%
Sorting networks are a kind of sorting algorithm that consist of wires carrying the values and comparators, which swap the values on two wires if they are not in the desired order. 
Sorting networks can be made differentiable by perturbing the values on the wires in each layer of the sorting network by a logistic distribution, i.e., instead of $\min$ and $\max$ they use $\operatorname{softmin}$ and $\operatorname{softmax}$.}
Similar to the methods above, this method produces a relaxed permutation matrix, which allows us to apply it to top-$k$ classification learning.
The method has also been improved by enforcing monotonicity and bounding the approximation error~\cite{petersen2022monotonic}.
\revA{%
Note that sorting networks are a classic algorithmic concept~\citep{Knuth1998-3-SortingSearching}, are not neural networks nor refer to differentiable sorting.
Differentiable sorting networks are one of multiple differentiable sorting and ranking methods.
}

\section{Top-$k$ Learning}

In this section, we start by introducing our objective, elaborate its exact formulation, and then build on differentiable sorting principles to efficiently approximate the objective.
A visual overview over the loss architecture is also given in Figure~\ref{fig:panda-overview}.

The goal of top-$k$ learning is to extend the learning criterion from only accepting exact (top-$1$) predictions to accepting $k$ predictions among which the correct class has to be.
In its general form, for top-$k$ learning, $k$ may differ for each application, class, data point, or a combination thereof. 
For example, in one case one may want to rank $5$ predictions and assign a score that depends on the rank of the true class among these ranked predictions, while, in another case, one may want to obtain $5$ predictions but does not care about their order.
In yet another case, such as image classification, one may want to enforce a top-$1$ accuracy on images from the ``person'' super-class, but resign to a top-$3$ accuracy for the ``animal'' super-class, as it may have more ambiguities in class-labels.
(For example, as recently shown by \citet{northcutt2021pervasive}, there is noise in the labels of ImageNet-1K. 
As ImageNet21K is a superset of ImageNet1K, it also holds in this case, with the addition that labeling in case of 21K classes would be more challenging and therefore more error-prone.)
We model this by a random variable~$K$, following a distribution $P_K$ that describes the relative importance of different values~$k$. 
The discrete distribution $P_K$ is either a marginalized distribution for a given setting (such as the uniform distribution), or a conditional distribution for each class, data point, etc. 
This allows specifying a marginalized or conditional distribution $k\sim P_K$. 
This generalizes the ideas of conventional top-$1$ supervision (usually softmax cross-entropy) and top-$k$ supervision for a $k$ like $k=5$ (usually based on surrogate top-$k$ margin/hinge losses like \citep{lapin2016loss, berrada2018smooth}) and unifies them.

The objective of top-$k$ learning is maximizing the probability of accepted predictions of the model $f_\Theta$ on data $X, y \sim \mathcal{D}$ given marginal distribution $P_K$ (or conditional $P_{K|X,y}$ if it depends on the class $y$ and/or data point $X$). In the following, $\mP_{k,y}$ is the predicted probability of $y$ being the $k$th-best prediction for data point $X$.
\begin{equation}
    \argmax_\Theta\ \ \mathbb{E}_{X, y \sim \mathcal{D}} \left[\log\left(
    \mathbb{E}_{k \sim P_K} \left[    
            {\textstyle\sum_{m=1}^k}
            \mP_{m, y}
    \right]\right)
    \right]
\end{equation}
To evaluate the probability of $y$ to be the top-$1$ prediction, we can simply use $\softmax_y(f_\Theta (X))$.
However, $k>1$ requires more consideration. 
Here, we require probability scores $\mP_{k, c}$ for the $k$th prediction over classes $c\in\mathbb{C}$, where $\sum_{c=1}^n \mP_{k, c} = 1$ (i.e., $\mP$ is row stochastic) and ideally additionally $\sum_{k=1}^n \mP_{k, c} = 1$ (i.e., $\mP$ is also column stochastic and thus doubly stochastic.)
With this, we can optimize our model by minimizing the following loss
\begin{equation}
    \mathcal{L}(X, y)\,{=}\,{-} \log\!\left( \sum_{k=1}^n P_K(k) \!\left(\sum_{m=1}^k \mP_{m, y}(f_\Theta (X)) \right)\!\!\right)\!\!
    \label{eq:top-k-learning-loss}
\end{equation}
which is the cross entropy over the probabilities that the true class is among the top-$k$ class for each possible~$k$. 
Note that $\sum_{k=1}^n P_K(k) = 1$. 

If $\bf{P}$ is column stochastic, the inner sum in Equation~\ref{eq:top-k-learning-loss} is $\leq1$. 
As the sum over $P_K$ is $1$, the outer sum is also $\leq1$.
$\bf{P}$ being column stochastic is the desirable case. This is given for DiffSortNets and SinkhornSort. However, for SoftSort and NeuralSort, this is only approximately the case.
In the non-column stochastic case of SoftSort and NeuralSort, the inner sum could become greater than $1$; however, we did not observe this to be a direct problem.

To compute $\mP_{k, c}$, we require a function mapping from a vector of real-valued scores to an (ideally) doubly stochastic matrix $\mP$.
The most suitable for this are the differentiable relaxations of the sorting and ranking functions, which produce differentiable permutation matrices $\mP$, which we introduced in Section~\ref{sec:background-diff-sort-and-rank}.
We build on these approximations to propose instances of top-$k$ learning losses and extend differentiable sorting networks to differentiable top-$k$ networks, as just finding the top-$k$ scores is computationally cheaper than sorting all elements and reduces the approximation error.

\subsection{Top-$k$ Probability Matrices}

The discussed differentiable sorting algorithms produce relaxed permutation matrices of size $n\times n$.
However, for top-$k$ classification learning, we require only the top $k$ rows for the number $k$ of top-ranked classes to consider.
\revC{Here, $k$ is the largest $k$ that is considered for the objective, i.e., where $P_K(k)>0$.}
As $n\gg k$, producing a $k\times n$ matrix instead of a $n\times n$ matrix is much faster.

For \textit{NeuralSort and SoftSort}, it is possible to simply compute only the top rows, as the algorithm is defined row-wise. 

For the \textit{differentiable Sinkhorn sorting algorithm}, it is not directly possible to improve the runtime, as in each Sinkhorn iteration the full matrix is required. 
\citet{xie2020differentiable} proposed a Sinkhorn-based differentiable top-$k$ operator, which
computes a $2\times n$ matrix where the first row corresponds to the top-$k$ elements and the second row correspond to the remaining elements. However, this formulation does not produce $\bf{P}$ and does not distinguish between the placements
of the top-$k$ elements among each other, and thus we use the SinkhornSort algorithm by \citet{Cuturi2019-SortingOT}.

For \textit{differentiable sorting networks}, it is (via a bi-directional evaluation) possible to reduce the cost from $\mathcal{O}(n^2 \log^2 (n))$ to $\mathcal{O}(nk \log^2 (n))$. 
Here, it is important to note the shape and order of multiplications for obtaining $P$. 
As we only need those elements, which are (after the last layer of the sorting network) at the top $k$ ranks that we want to consider, we can omit all remaining rows of the permutation matrix of the last layer (layer $t$) and thus it is only of size $(k\times n)$.
\begin{equation}
    \underbrace{(k\times n)}_{\text{$P$}} \  = \ \underbrace{(k\times n)}_{\text{layer $t$}}\ \underbrace{(n\times n)}_{\text{layer $t{-}1$}}\ \ ...\ \ \underbrace{(n\times n)}_{\text{layer $1$}}
    \label{eq:shapes-diffsort-topk}
\end{equation}
Note that during execution of the sorting network, $P$ is conventionally computed from layer $1$ to layer $t$, i.e., from right to left.
If we computed it in this order, we would only save a tiny fraction of the computational cost and only during the last layer.
Thus, we propose to execute the differentiable sorting network, save the values that populate the (sparse) $n\times n$ layer-wise permutation matrices, and compute $P$ in a second pass from the back to the front, i.e., from layer $t$ to layer $1$, or from left to right in Equation~\ref{eq:shapes-diffsort-topk}.
This allows executing $t$ dense-sparse matrix multiplications with dense $k\times n$ matrices and sparse $n\times n$ matrices instead of dense $n\times n$ and sparse $n\times n$ matrices. 
With this, we reduce the asymptotic complexity from $\mathcal{O}(n^2 \log^2 (n))$ to $\mathcal{O}(nk \log^2 (n))$.

\subsection{Differentiable Top-$k$ Networks}

As only the top-$k$ rows of a relaxed permutation matrix are required for top-$k$ classification learning, it is possible to improve the efficiency of computing the top-$k$ probability distribution via differentiable sorting networks by reducing the number of differentiable layers and comparators.
Thus, we propose differentiable top-$k$ networks, which relax selection networks in analogy to how differentiable sorting networks relax sorting networks.
Selection networks are networks that select only the top-$k$ out of $n$ elements \citep{Knuth1998-3-SortingSearching}.
We propose splitter selection networks (SSN), a novel class of selection networks that requires only $\mathcal{O}(\log n)$ layers (instead of the $\mathcal{O}(\log^2 n)$ layers for sorting networks) which makes top-$k$ supervision with differentiable top-$k$ networks more efficient and reduces the error (which is introduced in each layer.)
SSNs follow the idea that the input is split into locally sorted sublists and then all wires that are not candidates to be among the global top-$k$ can be eliminated.
For example, for $n=1024, k=5$, SSNs require only $22$ layers, while the best previous selection network requires $34$ layers and full sorting (with a bitonic network) requires even $55$ layers. 
For $n=10450, k=5$ (i.e., for ImageNet-21K-P), SNNs require $27$ layers, the best previous requires $50$ layers, and full sorting requires $105$ layers.
In addition, the layers of SSNs are less computationally expensive than those of the bitonic sorting network.
Details on SSNs, as well as their full construction, can be found in Supplementary Material~\ref{apx:ssn}.
Concluding, the contribution of differentiable top-$k$ networks is two-fold: first, we propose a novel kind of selection networks that needs fewer layers, and second, we relax those similarly to differentiable sorting networks.

\subsection{Implementation Details}
\label{sec:implementation_details}

Despite those performance improvements, evaluating the differentiable ranking operators still requires a considerable amount of computational effort for large numbers of classes. 
Especially if the number $n$ of elements to be ranked is $n=1\,000$ (ImageNet-1K) or even $n>10\,000$ (ImageNet-21K-P), the differentiable ranking operators can dominate the overall computational costs.
In addition, for large numbers $n$ of elements to be ranked, the performance of differentiable ranking operators decreases as differentially ranking more elements naturally introduces larger errors \citep{Grover2019-NeuralSort,prillo2020softsort,Cuturi2019-SortingOT,Petersen2021-diffsort}.
Thus, we reduce the number of outputs to be ranked differentially by only considering those classes (for each input) that have a score among the top-$m$ scores.
For this, we make sure that the ground truth class is among those top-$m$ scores, by replacing the lowest of the top-$m$ scores by the ground truth class, if necessary.
For $n=1000$, we choose $m=16$, and for $n>10\,000$, we choose $m=50$.
We find that this greatly improves training performance.

Because the differentiable ranking operators are (by their nature of being differentiable) only approximations to the hard ranking operator, they each have their characteristics and inconsistencies.
Thus, for training models from scratch, we replace the top-$1$ component of the loss by the regular softmax, which has a better and more consistent behavior.
This guides the other loss if the differentiable ranking operator behaves inconsistently.
To avoid the top-$k$ components affecting the guiding softmax component and avoid probabilities greater than $1$ in $p$, we can separate the cross-entropy into a mixture of the softmax cross-entropy ($\mathrm{sm}$, for the top-$1$ component) and the top-$k$ cross-entropy ($\mathrm{top}-k$, for the top-$k\geq2$ components) as follows: %

\begin{align}
    &\underset{\mathrm{sm+top-}k}{\mathcal{L}(X, y)}
    = P_K(1) \cdot  \operatorname{SoftmaxCELoss}(f_\Theta (X), y) \\[-0.7ex]
    &{-} (1-P_K(1)) \cdot \log\!\left( \sum_{k=2}^n P_K(k) \left(\sum_{m=1}^k \mP_{m, y}(f_\Theta (X)) \right)\!\!\right) \notag %
\end{align}

\section{Related Work}

We structure the related work into three broad sections: works that derive and apply differentiable top-$k$ operators, works that use ranking and top-$k$ training objectives in general, and works that present classic selection networks.

\subsection{Differentiable Top-$k$ Operators}

\citet{Grover2019-NeuralSort} include an experiment where they use the NeuralSort differentiable top-$k$ operator for $k$NN learning.
\citet{Cuturi2019-SortingOT}, \citet{Blondel2020-FastSorting}, and \citet{Petersen2021-diffsort} each apply their differentiable sorting and ranking methods to top-$k$ supervision with $k=1$.

\citet{xie2020differentiable} propose a differentiable top-$k$ operator based on optimal transport and the Sinkhorn algorithm \citep{Cuturi13Sinkhorn}.
They apply their method to $k$-nearest-neighbor learning ($k$NN), differential beam search with sorted soft top-$k$, and top-$k$ attention for machine translation.
\citet{cordonnier2021differentiable} use perturbed optimizers \citep{Berthet2020-PerturbedOptimizers} to derive a differentiable top-$k$ operator, which they use for differentiable image patch selection.
\citet{lee2021differentiable} propose using NeuralSort for a differentiable top-$k$ operator to produce differentiable ranking metrics for recommender systems.
\citet{goyal2018continuous} propose a continuous top-$k$ operator for differentiable beam search.
\citet{pietruszka2020successive} propose the differentiable successive halving top-$k$ operator to approximate the normalized Chamfer Cosine Similarity ($nCCS@k$).

\subsection{Ranking and Top-$k$ Training Objectives}

\citet{fan2017learning} propose the ``average top-$k$'' loss, an aggregate loss that averages over the $k$ largest individual losses of a training data set.
They apply this aggregate loss to SVMs for classification tasks.
Note that this is not a differentiable top-$k$ loss in the sense of this work. 
Instead, the top-$k$ is not differentiable and used for deciding which data points' losses are aggregated into the loss.

\citet{lapin2015top,lapin2016loss} propose relaxed top-$k$ surrogate error functions for multiclass SVMs.
Inspired by learning-to-rank losses, they propose top-$k$ calibration, a top-$k$ hinge loss, a top-$k$ entropy loss, as well as a truncated top-$k$ entropy loss.
They apply their method to multiclass SVMs and learn via stochastic dual coordinate ascent (SDCA).

\citet{berrada2018smooth} build on these ideas and propose smooth loss functions for deep top-$k$ classification.
Their surrogate top-$k$ loss achieves good performance on the CIFAR-100 and ImageNet1K tasks.
While their method does not improve performance on the raw data sets in comparison to the strong Softmax Cross-Entropy baseline, in settings of label noise and data set subsets, they improve classification accuracy.
Specifically, with label noise of $20\%$ or more on CIFAR-100, they improve top-$1$ and top-$5$ accuracy and for subsets of ImageNet1K of up to $50\%$ they improve top-$5$ accuracy.
This work is closest to ours in the sense that our goal is to improve learning of neural networks.
However, in contrast to \citep{berrada2018smooth}, our method improves classification accuracy in unmodified settings.
In our experiments, for the special case of $k$ being a concrete integer and not being drawn from a distribution, we provide comparisons to the smooth top-$k$ surrogate loss.

\citet{yang2020consistency} provide a theoretical analysis of top-$k$ surrogate losses 
as well as produce a new surrogate top-$k$ loss, which they evaluate in synthetic data experiments.

\revD{A related idea is set-valued classification, where a set of labels is predicted. We refer to \citet{chzhen2021set} for an extensive overview. We note that our goal is not to predict a set of labels, but instead we return a score for each class corresponding to a ranking, where only one class can correspond to the ground truth.}

\subsection{Selection Networks}

Previous selection networks have been proposed by, i.a., \citep{Wah_and_Chen_1984,Zazon-Ivry_and_Codish_2012,Karpinski_and_Piotrow_2015}.
All of these are based on classic divide-and-conquer sorting networks, which recursively sort subsequences and merge them. 
In selection networks, during merging, only the top-$k$ elements are merged instead of the full (sorted) subsequences.
In comparison to those earlier works, we propose a new class of selection networks, which achieve tighter bounds (for $k\ll n$), and relax them.

\section{Experiments\protect\footnote{Code will be available at~\href{https://github.com/Felix-Petersen/difftopk}{github.com/Felix-Petersen/difftopk}}}

\subsection{Setup}
We evaluate the proposed top-$k$ classification loss for four differentiable ranking operators on CIFAR-100, ImageNet-1K, as well as the winter 2021 edition of ImageNet-21K-P.
We use CIFAR-100, which can be considered a small-scale data set with only $100$ classes, to train a ResNet18 model \citep{he2016deep_resnet} from scratch and show the impact of the proposed loss function on the top-$1$ and top-$5$ accuracy.
In comparison, ImageNet-1K and ImageNet-21K-P provide rather large-scale data sets with $1\,000$ and $10\,450$ classes, respectively.
To avoid the unreasonable carbon-footprint of training many models from scratch, we decided to exclusively use publicly available backbones for all ImageNet experiments.
This has the additional benefit of allowing more settings, making our work easily reproducible, and allowing to perform multiple runs on different seeds to improve the statistical significance of the results.
For ImageNet-1K, we use two publicly available state-of-the-art architectures as backbones: 
First, the (four) ResNeXt-101 WSL architectures by \citet{mahajan2018exploring}, which were pretrained in a weakly-supervised fashion on a billion-scale data set from Instagram.
Second, the Noisy Student EfficientNet-L2 \citep{xie2020self}, which was pretrained on the unlabeled JFT-300M data set~\citep{sun2017revisiting}.
For ResNeXt-101 WSL, we extract $2\,048$-dimensional embeddings and for the Noisy Student EfficientNet-L2, we extract $5\,504$-dimensional embeddings of ImageNet-1K and fine-tune on them.

We apply the proposed loss in combination with various available differentiable sorting and ranking approaches, namely NeuralSort, SoftSort, SinkhornSort, and DiffSortNets. 
To determine the optimal temperature for each differentiable sorting method, we perform a grid search at a resolution of factor $2$. 
For training, we use the Adam optimizer \citep{Kingma2014AdamOpt}.
For training on CIFAR-100 from scratch, we train for up to $200$ epochs with a batch size of $100$ at a learning rate of $10^{-3}$.
For ImageNet-1K, we train for up to $100$ epochs at a batch size of $500$ and a learning rate of $10^{-4.5}$.
For ImageNet-21K-P, we train for up to $40$ epochs at a batch size of $500$ and a learning rate of $10^{-4}$.
We use early stopping and found that these settings lead to convergence in all settings.
As baselines, we use the respective original models, softmax cross-entropy, as well as learning with the smooth surrogate top-$k$ loss~\citep{berrada2018smooth}.

\begin{table}[t]
    \centering
    \addtolength{\tabcolsep}{-4pt}  
    \resizebox{.79\linewidth}{!}{
    \begin{tabular}{lcc}
\toprule
Method & $P_K$  &  {CIFAR-100} \\
\midrule
\underline{\textit{Baselines}}\\
Softmax                     & $([1,0,0,0,0])$     & $61.27\,|\,85.31$ \\[.2em]
Smooth\,top-$k$ loss $(*)$  & $([0,0,0,0,1])$     & $53.07\,|\,85.23$ \\
Top-$5$ NeuralSort          & $[0,0,0,0,1]$     & $22.58\,|\,84.41$ \\
Top-$5$ SoftSort            & $[0,0,0,0,1]$     & $1.01\,|\,5.09$   \\
Top-$5$ SinkhornSort        & $[0,0,0,0,1]$     & $55.62\,|\,\pmb{87.04}$ \\
Top-$5$ DiffSortNets        & $[0,0,0,0,1]$     & $52.81\,|\,84.21$ \\
\midrule
\underline{\textit{Ours}}\\
Top-$k$ NeuralSort          & $[.2,.2,.2,.2,.2]$   & $61.46\,|\,86.03$\\   %
Top-$k$ SoftSort            & $[.2,.2,.2,.2,.2]$   & $61.53\,|\,82.39$\\   %
Top-$k$ SinkhornSort        & $[.2,.2,.2,.2,.2]$   & $61.89\,|\,\pmb{86.94}$\\   %
Top-$k$ DiffSortNets        & $[.2,.2,.2,.2,.2]$   & $\pmb{62.00}\,|\,86.73$\\   %
\bottomrule
    \end{tabular}
    }
    \addtolength{\tabcolsep}{4pt}  
    \vspace*{-.25em}
    \caption{
        CIFAR-100 results for training a ResNet18 from scratch.
        The metrics are Top-$1\,|\,$Top-$5$ accuracy averaged over 2 seeds.
        $(*)$:~\citet{berrada2018smooth}.
    }
    \label{tab:cifar100-main}
    \vspace*{-2em}
\end{table}

\subsection{Training from Scratch}

We start by demonstrating that the proposed loss can be used to train a network from scratch. 
As a reference baseline, we train a ResNet18 from scratch on CIFAR-100.
In Table~\ref{tab:cifar100-main}, we compare the baselines (i.e., top-$1$ softmax, the smooth top-$5$ loss \citep{berrada2018smooth}, as well as ``pure'' top-$5$ losses using four differentiable sorting and ranking methods) with our top-$k$ loss with $k\sim[.2,.2,.2,.2,.2]$.

We find that training with top-$5$ alone|in some cases|slightly improves the top-$5$ but has a substantially worse top-$1$ accuracy.
Here, we note that the smooth top-$5$ loss \citep{berrada2018smooth}, top-$5$ Sinkhorn \citep{Cuturi2019-SortingOT}, and top-$5$ DiffSort \citep{Petersen2021-diffsort} are able to achieve good performance.
Notably, Sinkhorn \citep{Cuturi2019-SortingOT} outperforms the softmax baseline on the top-$5$ metric, while NeuralSort and SoftSort are less stable and yield worse results especially on top-$1$ accuracy. %

By using our loss that corresponds to drawing $k$ from $[.2,.2,.2,.2,.2]$, we can achieve substantially improved results, especially also on the top-$1$ accuracy metric.
Using the DiffSortNets yields the best results on the top-$1$ accuracy and Sinkhorn yields the best results on the top-$5$ accuracy.
Note that, here, also NeuralSort and SoftSort achieve good results in this setting, which can be attributed to our loss with $k\sim[.2,.2,.2,.2,.2]$ being more robust to inconsistencies and outliers in the used differentiable sorting method.
Interestingly, top-$5$ SinkhornSort achieves the best performance on the top-$5$ metric, which suggests that SinkhornSort is a very robust differentiable sorting method as it does not require additional top-$k$ components.
Nevertheless, it is advisable to include other top-$k$ components as the model trained purely on top-$5$ exhibits poor top-$1$ performance.

\begin{table}[t]
    \centering
    \addtolength{\tabcolsep}{-4pt}  
    \resizebox{\linewidth}{!}{
    \begin{tabular}{lccc}
\toprule
Method & $P_K$  &  {ImgNet-1K} & \kern-.25emImgNet-21K-P\\
\midrule
\underline{\textit{Baselines}}\\
Softmax                     & $([1,0,0,0,0])$     & $86.06\,|\,97.795$ & $39.29\,|\,69.63$ \\[.2em]
Smooth\,top-$k$ loss $(*)$  & $([0,0,0,0,1])$     & $85.15\,|\,97.540$ & $34.03\,|\,65.56$ \\
Top-$5$ NeuralSort          & $[0,0,0,0,1]$     & $33.37\,|\,94.748$ & $15.87\,|\,33.81$ \\
Top-$5$ SoftSort            & $[0,0,0,0,1]$     & $18.23\,|\,94.965$ & $33.61\,|\,69.82$ \\
Top-$5$ SinkhornSort        & $[0,0,0,0,1]$     & $85.65\,|\,\pmb{97.991}$ & $36.93\,|\,69.80$ \\
Top-$5$ DiffSortNets        & $[0,0,0,0,1]$     & $69.05\,|\,97.389$ & $35.96\,|\,69.76$ \\
\midrule
\underline{\textit{Ours}}\\
Top-$k$ NeuralSort          & $[.5,0,0,0,.5]$   & $86.30\,|\,97.896$        &  $37.85\,|\,68.08$ \\
Top-$k$ SoftSort            & $[.5,0,0,0,.5]$   & $86.26\,|\,97.963$        &  $39.93\,|\,70.63$ \\
Top-$k$ SinkhornSort        & $[.5,0,0,0,.5]$   & $\pmb{86.29}\,|\,\pmb{97.971}$  &  $39.85\,|\,70.56$ \\
Top-$k$ DiffSortNets        & $[.5,0,0,0,.5]$   & $86.24\,|\,97.937$        &  \pmb{$40.22\,|\,70.88$} \\
\bottomrule
    \end{tabular}
    }
    \addtolength{\tabcolsep}{4pt}  
    \vspace*{-.75em}
    \caption{
        ImageNet-1K and ImageNet-21K-P results for fine-tuning the head of ResNeXt-101 32x48d WSL \citep{mahajan2018exploring}.
        The metrics are Top-$1\,|\,$Top-$5$ accuracy averaged over 10 seeds for ImageNet-1K and 2 seeds for ImageNet-21K-P.
        $(*)$:~\citet{berrada2018smooth}.
    }
    \label{tab:imagenet-main}
    \vspace*{-2em}
\end{table}

\begin{figure*}
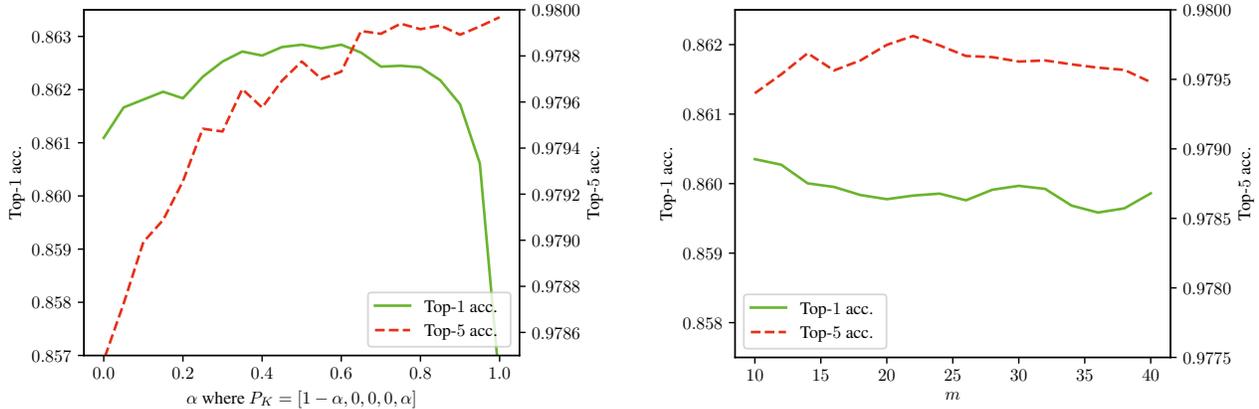

    \centering
    \resizebox{.49\textwidth}{!}{\input{fig/imagenet1k-top1-vs-top5-loss-plot-sinkhorn.pgf}}\hfill%
    \resizebox{.49\textwidth}{!}{\input{fig/imagenet1k-varying-m-plot-sinkhorn_0_75.pgf}}
    \vspace{-1.5em}
    \caption{
    Effects of varying the ratio between top-$1$ and top-$5$ (left) and varying the size of differentially ranked subset $m$.
    Both experiments are done with the differentiable Sinkhorn ranking algorithm \citep{Cuturi2019-SortingOT}.
    On the left, $m=16$, on the right, $\alpha=0.75$.
    Averaged over $5$ runs.
    }
    \label{fig:top1-vs-top5-loss-plot}
    \label{fig:different-ms-plot}
    \vspace*{-1.5em}
\end{figure*}

\begin{figure}[t]
    \centering
    \resizebox{\linewidth}{!}{\input{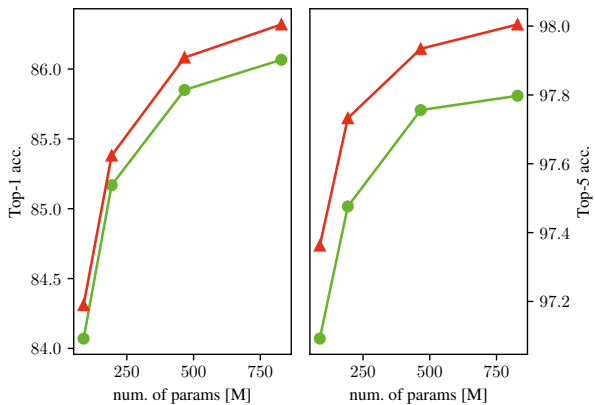}}
    \vspace{-2em}
    \caption{
        ImageNet-1K accuracy improvements for all ResNeXt-101 WSL model sizes (32x8d, 32x16d, 32x32d, 32x48d). %
        Green ($\bullet$) is the original model and red ($\blacktriangle$) is with top-$k$ fine-tuning.
        }
    \label{fig:resnext-wsl-different-model-sizes-improvements-plot}
    \vspace*{-1em}
\end{figure}

\subsection{Fine-Tuning}

In this section, we discuss the results for fine-tuning on ImageNet-1K and ImageNet-21K-P.
In Table~\ref{tab:imagenet-main}, we find a very similar behavior to training from scratch on CIFAR-100.
Specifically, we find that training accuracies improve by drawing $k$ from a distribution. 
An exception is (again) SinkhornSort, where focussing only on top-$5$ yields the best top-$5$ accuracy on ImageNet-1K, but the respective model exhibits poor top-$1$ accuracy.
Overall, we find that drawing $k$ from a distribution improves performance in all cases.

To demonstrate that the improvements also translate to different backbones, we show the improvements on all four model sizes of ResNeXt-101 WSL (32x8d, 32x16d, 32x32d, 32x48d) in Figure~\ref{fig:resnext-wsl-different-model-sizes-improvements-plot}.
Also, here, our method improves the model in all settings.

\subsection{Impact of the Distribution $P_K$ and Differentiable Sorting Methods}

We start by demonstrating the impact of $P_K$, which is the distribution from which we draw $k$.
Let us first consider the case where $k$ is $5$ with probability $\alpha$ and $1$ with probability $1-\alpha$, i.e., $P_K=[1-\alpha, 0, 0, 0, \alpha]$.
In Figure~\ref{fig:top1-vs-top5-loss-plot} (left), we demonstrate the impact that changing $\alpha$, i.e., transitioning from a pure top-$1$ loss to a pure top-$5$ loss, has on fine-tuning ResNeXt-101 WSL with our loss using the SinkhornSort algorithm.
Increasing the weight of the top-$5$ component does not only increase the top-$5$ accuracy but also improves the top-$1$ accuracy up to around $60\%$ top-$5$; when using only $k=5$, the top-$1$ accuracy drastically decays as the incentive for the true class to be at the top-$1$ position vanishes (or is only indirectly given by being among the top-$5$.)
While the top-$5$ accuracy in this plot is best for a pure top-$5$ loss, this generally only applies to the Sinkhorn algorithm and overall training is more stable if a pure top-$5$ is avoided. 
This can also be seen in Tables~\ref{tab:cifar100-main} and~\ref{tab:imagenet-main}.

In Tables~\ref{tab:imagenet-different-PK} and \ref{tab:cifar100-different-PK}, we consider more additional settings with all differentiable ranking methods.
Specifically, we compare four notable settings:
$[.5,0,0,0,.5]$, i.e., equally weighted top-$1$ and top-$5$; $[.25,0,0,0,.75]$ and $[.1,0,0,0,.9]$, i.e., top-$5$ has larger weights; $[.2, .2, .2, .2, .2]$, i.e., the case of having an equal weight of $0.2$ for top-$1$ to top-$5$.
The $[.5,0,0,0,.5]$ setting is a rather canonical setting which usually performs well on both metrics, while the others tend to favor top-$5$.
In the $[.5,0,0,0,.5]$ setting, all sorting methods improve upon the softmax baseline on both top-$1$ and top-$5$ accuracy.
When increasing the weight of the top-$5$ component, the top-$5$ generally improves while top-$1$ decays.

Here we find a core insight of this paper: the best performance cannot be achieved by optimizing top-$k$ for only a single $k$, but instead, drawing $k$ from a distribution improves performance on all metrics.

Comparing the differentiable ranking methods, we can find the overall trend that SoftSort outperforms NeuralSort, and that SinkhornSort as well as DiffSortNets perform best.
We can see that some sorting algorithms are more sensitive to the overall $P_K$ than others:
Whereas SinkhornSort \citep{Cuturi2019-SortingOT} and DiffSortNets \citep{Petersen2021-diffsort} continuously outperform the softmax baseline, NeuralSort \citep{Grover2019-NeuralSort} and SoftSort \citep{prillo2020softsort} tend to collapse when over-weighting the top-$5$ components.

Comparing the performance on the medium-scale ImageNet-1K to the larger ImageNet-21K-P in Table~\ref{tab:imagenet-main}, we observe a similar pattern. 
Here, again, using the top-$k$ component alone is not enough to significantly increase accuracy, but combining top-$1$ and top-$k$ components helps to improve accuracy on both reported metrics. 
While NeuralSort struggles in this large-scale ranking problem and stays below the softmax baseline, DiffSortNets~\citep{Petersen2021-diffsort} provide the best top-$1$ and top-$5$ accuracy with $40.22\%$ and $70.88\%$, respectively.

In Supplementary Material~\ref{apx:extension-10-20}, an extension to learning with top-$10$ and top-$20$ components can be found. 

We note that we do not claim that all settings (especially all differentiable sorting methods) improve the classification performance on all metrics. 
Instead, we include all methods and also additional settings to demonstrate the capabilities and limitations of each differentiable sorting method.

Overall, it is notable that SinkhornSort achieves the overall most robust training behavior, while also being by far the slowest sorting method and thus potentially slowing down training drastically, especially when the task is only fine-tuning.
SinkhornSort tends to require more Sinkhorn iterations towards the end of training.
DiffSortNets are considerably faster, especially, it is possible to only compute the top-$k$ probability matrices and because of our advances for more efficient selection networks.

\begin{table}[t]
    \centering
    \addtolength{\tabcolsep}{-4pt}  
    \resizebox{\linewidth}{!}{
    \begin{tabular}{lcccc}
\toprule
Method~/~$P_K$    & {\small$[.5,0,0,0,.5]$} & {\small$[.25,0,0,0,.75]$} & {\small$[.1,0,0,0,.9]$} & {\small$[.2,.2,.2,.2,.2]$}\\
\midrule
\textit{ImageNet-1K}\\
NeuralSort          & $86.30\,|\,97.896$  &  $34.26\,|\,95.410$  &  $34.32\,|\,94.889$  &  $85.75\,|\,97.865$\\
SoftSort            & $86.26\,|\,97.963$  &  $86.16\,|\,97.954$  &  $27.30\,|\,95.915$  &  $86.18\,|\,97.979$\\
SinkhornSort        & $\pmb{86.29}\,|\,97.971$  &  $86.24\,|\,97.989$  &  $86.18\,|\,97.987$  &  $86.22\,|\,97.989$\\  %
DiffSortNets        & $86.24\,|\,97.937$  &  $86.15\,|\,97.936$  &  $86.04\,|\,97.980$  &  $86.21\,|\,\pmb{98.003}$\\  %
\midrule
\textit{ImageNet-21K-P}\kern-5em\\
NeuralSort          & $37.85\,|\,68.08$  &  $36.16\,|\,67.60$  &  $33.02\,|\,67.29$  &  $37.09\,|\,67.90$\\
SoftSort            & $39.93\,|\,70.63$  &  $39.08\,|\,70.27$  &  $37.78\,|\,70.07$  &  $39.68\,|\,70.57$\\
SinkhornSort        & $39.85\,|\,70.56$  &  $39.21\,|\,70.41$  &  $38.42\,|\,70.12$  &  $39.22\,|\,70.49$\\
DiffSortNets        & \pmb{$40.22\,|\,70.88$}  &  $39.56\,|\,70.58$  &  $38.48\,|\,70.25$  &  $39.69\,|\,70.69$\\  %
\bottomrule
    \end{tabular}
    }
    \addtolength{\tabcolsep}{4pt}  
    \vspace*{-.5em}
    \caption{
        ImageNet-1K and ImageNet-21K-P results for different distributions $P_K$ for fine-tuning the head of ResNeXt-101 32x48d WSL \citep{mahajan2018exploring}.
        The metrics are Top-$1\,|\,$Top-$5$ accuracy averaged over 10 seeds for ImageNet-1K and 2 seeds for ImageNet-21K-P.
    }
    \label{tab:imagenet-different-PK}
    \vspace*{-1em}
\end{table}

\subsection{Differentiable Ranking Set Size $m$}
We consider how accuracy is affected by varying the number of scores $m$ to be differentially ranked. 
Generally, the runtime of differentiable top-$k$ operators depends between linearly and cubic on $m$; thus it is important to choose an adequate value for $m$.
The choice of $m$ between $10$ and $40$ has only a moderate impact on the accuracy as can be seen in Figure~\ref{fig:different-ms-plot} (right).
However, when setting $m$ to large values such as $1\,000$ or larger, we observe that the differentiable sorting methods tend to become unstable.
We note that we did not specifically tune $m$, and that better performance can be achieved by fine-tuning $m$, as displayed in the plot.

\begin{table}[t]
    \centering
    \addtolength{\tabcolsep}{-4pt}  
    \resizebox{\linewidth}{!}{
    \begin{tabular}{lcccc}
\toprule
Method~/~$P_K$    & {\small$[.5,0,0,0,.5]$} & {\small$[.25,0,0,0,.75]$} & {\small$[.1,0,0,0,.9]$} & {\small$[.2,.2,.2,.2,.2]$}\\
\midrule
\textit{CIFAR-100}\kern-5em\\
NeuralSort          &  $61.12\,|\,86.47$  &  $61.07\,|\,87.23$  &  $52.57\,|\,85.76$  &  $61.46\,|\,86.03$\\   %
SoftSort            &  $61.17\,|\,83.95$  &  $61.05\,|\,83.10$  &  $58.16\,|\,79.26$  &  $61.53\,|\,82.39$\\   %
SinkhornSort        &  $61.34\,|\,86.38$  &  $61.50\,|\,86.68$  &  $57.35\,|\,86.34$  &  $61.89\,|\,86.94$\\   %
DiffSortNets        &  $60.07\,|\,86.44$  &  $61.57\,|\,86.51$  &  $61.74\,|\,\pmb{87.22}$  &  $\pmb{62.00}\,|\,86.73$\\   %
\bottomrule
    \end{tabular}
    }
    \addtolength{\tabcolsep}{4pt}  
    \vspace*{-.5em}
    \caption{
        CIFAR-100 results for different distributions $P_K$ for training a ResNet18 from scratch.
        The metrics are Top-$1\,|\,$Top-$5$ accuracy averaged over 2 seeds.
    }
    \label{tab:cifar100-different-PK}
    \vspace*{-.25em}
\end{table}

\begin{table}[t]
    \centering
    \resizebox{\linewidth}{!}{
    \begin{tabular}{llrcc}
    \toprule
        Method & & \kern-1.5em Public & Top-$1$ & Top-$5$  \\
    \midrule
        ResNet50 &$(*)$                                      & \cmark &  $79.26$  & $94.75 $ \\
        ResNet152 &$(*)$                                     & \cmark &  $80.62$  & $95.51 $ \\
        ResNeXt-101 32x48d WSL &$(\dagger)$              & \cmark & $85.43$	 & $97.57$ \\
        ViT-L/16 &$(\ddagger)$                            & \cmark & $87.76 $  & ---   \\
        Noisy Student EfficientNet-L2 &$(\mathsection)$                & \cmark & \pmb{$88.35$}   & \pmb{$98.65$} \\ 
        \midrule
        BiT-L   &$(\mathparagraph)$                                & \xmark & $87.54 $  & $98.46$  \\
        CLIP (w/ Noisy Student EffNet-L2) &$(\bigstar)$    & \xmark & $\approx88.4$    & --- \\ 
        ViT-H/14 &$(\divideontimes)$                            & \xmark & $88.55 $  & ---   \\
        ALIGN (EfficientNet-L2) &$(\doublebarwedge)$                   & \xmark & $88.64$   & \pmb{$98.67$} \\
        Meta Pseudo Labels (EffNet-L2)\kern-.5em &$(\doublecap)$        & \xmark & $90.20 $  & $\approx98.8$  \\
        ViT-G/14 &$(\doublecup)$                                 & \xmark & $90.45 $  & ---   \\
        CoAtNet-7 &$(\circ)$                                 & \xmark & \pmb{$90.88$}   & --- \\
    \midrule
        ResNeXt-101 32x48d WSL &&& $86.06 $ & $97.80 $ \\ 
        Top-$k$ SinkhornSort                                                &&& \pmb{$86.22 $} & $97.99 $ \\ %
        Top-$k$ DiffSortNets                                                &&& $86.21 $ & \pmb{$98.00 $} \\
    \midrule
        Noisy Student EfficientNet-L2  &&& $88.33 $ & $98.65 $ \\ 
        Top-$k$ SinkhornSort                                                &&& $88.32 $ & $98.66 $ \\
        Top-$k$ DiffSortNets                                                &&& \pmb{$88.37$} & \pmb{$98.68$} \\
    \bottomrule
    \end{tabular}
    }
    \vspace*{-.5em}
    \caption{
        ImageNet-1K result comparison to state-of-the-art.
        Among the overall best performing differentiable sorting / ranking methods, almost all results in reasonable settings outperform their respective baseline on Top-$1$ and Top-$5$ accuracy.
        For publicly available models / backbones, we achieve a new state-of-the-art for top-$1$ and top-$5$ accuracy. Our results are averaged over $10$ runs.
        $(*)$:~\citet{he2016deep_resnet}, 
        $(\dagger)$:~\citet{mahajan2018exploring}, 
        $(\ddagger)$:~\citet{dosovitskiy2021image}, 
        $(\mathsection)$:~\citet{xie2020self}, 
        $(\mathparagraph)$:~\citet{kolesnikov2020big}, 
        $(\bigstar)$:~\citet{radford2021learning}, 
        $(\divideontimes)$:~\citet{dosovitskiy2021image}, 
        $(\doublebarwedge)$:~\citet{jia2021scaling}, 
        $(\doublecap)$:~\citet{pham2021meta}, 
        $(\doublecup)$:~\citet{zhai2021scaling}, 
        $(\circ)$:~\citet{dai2021coatnet}%
        .
    }
    \label{tab:sota}
\end{table}

\subsection{Comparison to the State-of-the-Art}

We compare the proposed results to current state-of-the-art methods in Table~\ref{tab:sota}. 
We focus on methods that are publicly available and build upon two of the best performing models, namely Noisy Student EfficientNet-L2 \citep{xie2020self}, and ResNeXt-101 32x48d WSL \citep{mahajan2018exploring}.
Using both backbones, we achieve improvements on both metrics, and when fine-tuning on the Noisy Student EfficientNet-L2, we achieve a new state-of-the-art for publicly available models.

\paragraph{Significance Tests.} To evaluate the significance of the results, we perform a $t$-test (with significance level of $0.01$).
We find that our model is significantly better than the original model on both top-$1$ and top-$5$ accuracy metrics. 
Comparing to the observed accuracies of the baseline ($88.33\,|\,98.65$), DiffSortNets are significantly better (p=$0.00001\,|\,0.00005$). 
Comparing to the reported accuracies of the baseline ($88.35\,|\,98.65$), DiffSortNets are also significantly better (p=$0.00087\,|\,0.00005$).

\section{Conclusion}

We presented a novel loss, which relaxes the assumption of using a fixed $k$ for top-$k$ classification learning.
For this, we leveraged recent differentiable sorting and ranking operators.
We performed an array of experiments to explore different top-$k$ classification learning settings and achieved a state-of-the-art on ImageNet for publicly available models.

\subsection*{Acknowledgments \& Funding Disclosure}

This work was supported by 
the IBM-MIT Watson AI Lab, 
the DFG in the Cluster of Excellence EXC 2117 ``Centre for the Advanced Study of Collective Behaviour'' (Project-ID 390829875),
and the Land Salzburg within the WISS 2025 project IDA-Lab (20102-F1901166-KZP and 20204-WISS/225/197-2019).

\bibliography{manual}
\bibliographystyle{icml2022}

\newpage
\appendix

\section{Extension to Top-10 and Top-20}
\label{apx:extension-10-20}

We further extend the training settings, measuring the impact of top-$10$ and top-$20$ components on the large-scale ImageNet-21K-P dataset. 
The results are diplayed in Table~\ref{tab:imagenet-21k-p-top-10-top-20}, where we report top-$1$, top-$5$, top-$10$, and top-$20$ accuracy for all configurations. 
Again, we observe that $50\%$ top-$1$ and $50\%$ top-$k$ produces the overall best performance and that training with top-$5$ yields the best top-$1$, top-$5$, and top-$10$ accuracy.
We observe that the performance decays for top-$20$ components because (even among $10\,450$ classes) there are virtually no top-$20$ ambiguities, and artifacts of differentiable sorting methods can cause adverse effects.
Note that top-$10$ ambiguities do exist in ImageNet-21K-P, e.g., there are $11$ class hierarchy levels \cite{ridnik2021imagenet}.

\begin{table*}[h]
    \centering
    \resizebox{\linewidth}{!}{
    \begin{tabular}{lcccccccccccccccccccc}
\toprule
\textit{IN-21K-P}~~/~~$P_K$ $(@5)$\kern-1.5em  &  $\underbrace{[1,0,...]}_{\text{len}=5}$  &  $\underbrace{[.5,0,...,0,.5]}_{\text{len}=5}$  &  $\underbrace{[.25,0,...,0,.75]}_{\text{len}=5}$   &  $\underbrace{[.1,0,...,0,.9]}_{\text{len}=5}$ \\
\midrule
Softmax   (baseline)                         &  $39.29\,|\,69.63\,|\,78.55\,|\,85.33$\kern-4em  &  ---  &  ---  &  ---\\
NeuralSort      &  ---  &  $37.85\,|\,68.08\,|\,77.22\,|\,84.21$  &  $36.16\,|\,67.60\,|\,76.96\,|\,84.08$  &  $33.02\,|\,67.29\,|\,76.88\,|\,84.05$\\  
SoftSort           &  ---  &  $39.93\,|\,70.63\,|\,79.45\,|\,85.96$  &  $39.08\,|\,70.27\,|\,79.29\,|\,85.94$  &  $37.78\,|\,70.07\,|\,79.19\,|\,85.87$\\   
SinkhornSort     &  ---  &  $39.85\,|\,70.56\,|\,79.53\,|\,86.13$  &  $39.21\,|\,70.41\,|\,79.54\,|\,86.18$  &  $38.42\,|\,70.12\,|\,79.44\,|\,86.12$\\
DiffSortNets    &  ---  &  $\pmb{40.22}\,|\,\pmb{70.88}\,|\,\pmb{79.54}\,|\,86.03$  &  $39.56\,|\,70.58\,|\,79.44\,|\,86.01$  &  $38.48\,|\,70.25\,|\,79.29\,|\,85.90$\\ %
\bottomrule
\toprule
\textit{IN-21K-P}~~/~~$P_K$ $(@10)$\kern-1.5em  &  $\underbrace{[1,0,...]}_{\text{len}=10}$  &  $\underbrace{[.5,0,...,0,.5]}_{\text{len}=10}$  &  $\underbrace{[.25,0,...,0,.75]}_{\text{len}=10}$   &  $\underbrace{[.1,0,...,0,.9]}_{\text{len}=10}$ \\
\midrule
Softmax   (baseline)                         &  $39.33\,|\,69.62\,|\,78.55\,|\,85.36$\kern-4em  &  ---  &  ---  &  ---\\
NeuralSort      &  ---  &  $37.22\,|\,67.02\,|\,76.75\,|\,84.10$  &  $34.59\,|\,66.09\,|\,76.46\,|\,84.01$  &  $29.60\,|\,65.16\,|\,76.26\,|\,84.01$\\
SoftSort           &  ---  &  $39.26\,|\,69.52\,|\,79.13\,|\,85.93$  &  $37.71\,|\,68.56\,|\,78.71\,|\,85.78$  &  $33.68\,|\,67.35\,|\,78.43\,|\,85.70$\\
SinkhornSort     &  ---  &  $39.65\,|\,70.25\,|\,79.47\,|\,86.22$  &  $38.90\,|\,69.91\,|\,79.41\,|\,\pmb{86.25}$  &  $37.98\,|\,69.57\,|\,79.33\,|\,86.16$\\
DiffSortNets    &  ---  &  $39.92\,|\,70.13\,|\,79.38\,|\,86.02$  &  $39.10\,|\,69.60\,|\,79.21\,|\,86.03$  &  $37.88\,|\,69.07\,|\,79.04\,|\,85.91$\\
\bottomrule
\toprule
\textit{IN-21K-P}~~/~~$P_K$ $(@20)$\kern-1.5em  &  $\underbrace{[1,0,...]}_{\text{len}=20}$  &  $\underbrace{[.5,0,...,0,.5]}_{\text{len}=20}$  &  $\underbrace{[.25,0,...,0,.75]}_{\text{len}=20}$   &  $\underbrace{[.1,0,...,0,.9]}_{\text{len}=20}$ \\
\midrule
Softmax   (baseline)                         &  $39.33\,|\,69.62\,|\,78.55\,|\,85.36$\kern-4em  &  ---  &  ---  &  ---\\
NeuralSort      &  ---  &  $36.32\,|\,65.33\,|\,75.82\,|\,83.99$  &  $33.00\,|\,62.99\,|\,74.84\,|\,83.83$  &  $27.35\,|\,60.34\,|\,74.02\,|\,83.77$\\
SoftSort           &  ---  &  $38.04\,|\,65.98\,|\,77.17\,|\,85.45$  &  $34.30\,|\,62.89\,|\,76.03\,|\,85.19$  &  $24.02\,|\,56.35\,|\,74.32\,|\,84.82$\\
SinkhornSort     &  ---  &  $39.76\,|\,69.76\,|\,79.17\,|\,86.17$  &  $38.77\,|\,69.18\,|\,78.99\,|\,86.20$  &  $37.71\,|\,68.68\,|\,78.86\,|\,86.16$\\
DiffSortNets    &  ---  &  $39.54\,|\,68.59\,|\,77.95\,|\,85.49$  &  $38.62\,|\,67.62\,|\,77.43\,|\,85.37$  &  $37.46\,|\,66.80\,|\,77.01\,|\,85.17$\\
\bottomrule
    \end{tabular}
    }
    \caption{ImageNet 21K with top-$5$, top-$10$ and top-$20$ components.
    The displayed metrics per column are (Top-$1\,|\,$Top-$5\,|\,$Top-$10\,|\,$Top-$20$).
    }
    \label{tab:imagenet-21k-p-top-10-top-20}
\end{table*}

\section{Splitter Selection Networks}
\label{apx:ssn}

Similar to a sorting network, a \emph{selection network} is generally
a comparator network and hence it consists of wires (or lanes) carrying
values and comparators (or conditional swap devices) connecting pairs
of wires. A comparator swaps the values on the wires it connects if they
are not in a desired order. However, in contrast to a sorting network,
which sorts all the values carried by its wires, a $(k,n)$ selection
network, which has $n$ wires, moves the $k \le n$ largest (or, alternatively,
the $k$ smallest) values to a specific set of wires \citep{Knuth1998-3-SortingSearching},
most conveniently consecutive wires on one side of the wire array.
Note that the notion of a selection network usually does not require
that the selected values are sorted. However, in our context it is
preferable that they are, so that $P_K$ can easily be applied, and
the selection networks discussed below all have this property.

Clearly, any sorting network could be used as a selection network,
namely by focusing only on the top~$k$ (or bottom~$k$) wires. However,
especially if $k$ is small compared to~$n$, it is possible to construct
selection networks with smaller size (i.e.\ fewer comparators) and often
lower depth (i.e.\ a smaller number of layers, where a layer is a set of
comparators that can be executed in parallel).

A core idea of constructing selection networks was proposed in
\citep{Wah_and_Chen_1984}, based on the odd-even merge and bitonic sorting
networks \citep{Batcher_1968}: partition the $n$~wires into subsets of at
least~$k$ wires (preferably $2^{\lceil\log_2(k)\rceil}$ wires per subset)
and sort each subset with odd-even mergesort. Then merge the (sorted)
top~$k$ elements of each subsets with bitonic merge, thus halving the
number of (sorted) subsets. Repeat merging pairs of (sorted) subsets
until only a single (sorted) subset remains, the top $k$ elements of
which are the desired selection. This approach requires
$\frac{1}{2}\lceil\log_2(k)\rceil(\lceil\log_2(k)\rceil+1)
+(\lceil\log_2(n)\rceil-\lceil\log_2(k)\rceil)
 (\lceil\log_2(k)\rceil+1)$ layers.

Improvements to this basic scheme were developed in
\citep{Zazon-Ivry_and_Codish_2012,Karpinski_and_Piotrow_2015} and either rely
entirely on odd-even merge \citep{Batcher_1968} or entirely on pairwise
sorting networks \citep{Parberry_1992}. Especially selection networks based on
pairwise sorting networks have advantages in terms of the size of the
resulting network (i.e.\ number of needed comparators). However, these
improvements do not change the depth of the networks, that is, the
number of layers, which is most important in the context considered
here.

\begin{figure*}[t]
\centering
\includegraphics[width=.2\linewidth]{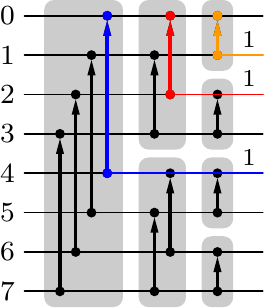}\hfill
\includegraphics[width=.2\linewidth]{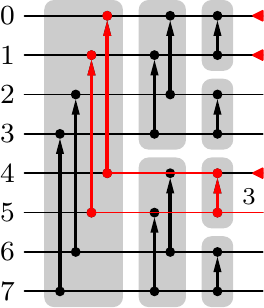}\hfill
\includegraphics[width=.2\linewidth]{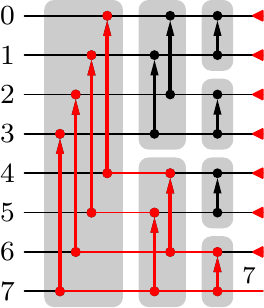}\hfill
\includegraphics[width=.2\linewidth]{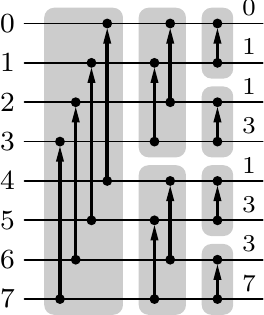}
\caption{\label{fig:splitter}Minimum ranks after a splitter cascade
  resulting from the transitive closure of the swaps.}
\end{figure*}

Our own selection network construction draws on this work by focussing
on a specific ingredient of pairwise sorting networks, namely a
so-called splitter (which happens to be identical to a single
bitonic merge layer, but for our purposes it is more comprehensible to
refer to it as a splitter). A~splitter for a list of $m$~wires having
indices~$[\ell_0,\ldots,\ell_{m-1}]$ has comparators connecting
wires~$\ell_i$ and $\ell_{i+s}$ where $s = \lceil\log_2(m)\rceil-1$
for $i \in \{ 0,\ldots, m-1 -2^{\lceil\log_2(m)\rceil)-1} \}$.

\begin{figure*}[t]
\centering
\includegraphics{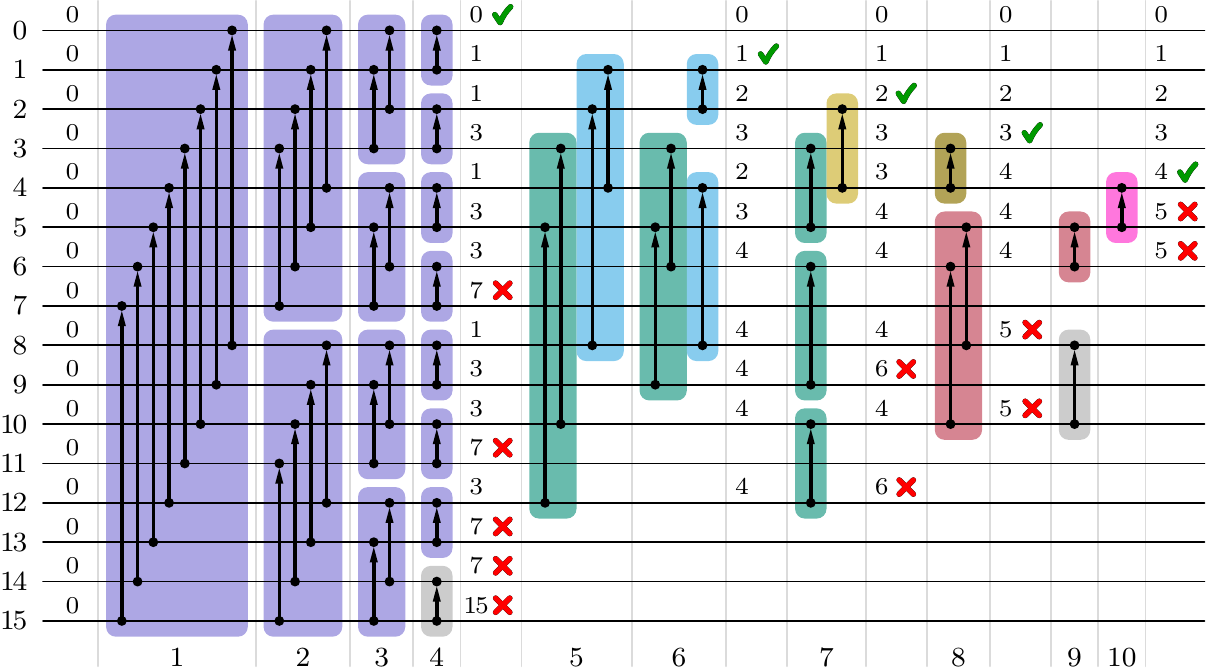}
\caption{\label{fig:selnet}A $(5,16)$ selection network constructed
  with the method described in the text. The numbers on the wires are
  the minimum ranks (starting at 0) that can be occupied by the values
  on these wires. Red crosses mark where wires can be excluded, green
  check marks where a top rank is determined. Swaps in blocks of equal
  color belong to the same splitter cascade. Swaps in gray boxes would
  be needed for full splitter cascades, but are not needed to determine
  the top 5 ranks.}
\end{figure*}

A pairwise sorting network starts with what we
call a {\em splitter cascade}. That is, an initial splitter
partitions the input wires into subsets of (roughly) equal size.
Each subset is split recursively until wire singletons result
\citep{Zazon-Ivry_and_Codish_2012}. An example of such a splitter cascade is
shown in Figure~\ref{fig:splitter} for 8~wires and in purple color for
16 wires in Figure~\ref{fig:selnet}
(arrows point to where the larger value is desired).

After a splitter cascade, the value carried by wire~$\ell_i$ has a
minimum rank of $r = 2^{b(i)}{-}1$, where $b(i)$ counts the number of
set bits in the binary number representation of~$i$. This minimum rank
results from the transitivity of the swap operations in the splitter
cascade, as is illustrated in Figure~\ref{fig:splitter} for 8~wires:
By following upward paths (in splitters to the left) through the
splitter cascade, one can find for each wire~$\ell_i$ exactly
$r = 2^{b(i)}{-}1$ wires with smaller indices that must carry values
no less than the value carried by wire~$\ell_i$. This yields the
minimum ranks shown in Figure~\ref{fig:splitter} on the right.

The core idea of our selection network construction is to use splitter
cascades to increase the minimum ranks of (the values carried by) wires.
If such a minimum rank exceeds~$k$ (or equals~$k$, since we work with
zero-based ranks and hence are interested in ranks
$\{0,\ldots,k{-}1\}$), a wire can be discarded, since its value is
certainly not among the top~$k$. On the other hand, if there is only
one wire with minimum rank~0, the top~1 value has been determined.
More generally, if all minimum ranks no greater than some value~$r$
occur for one wire only, the top~$r+1$ values have been determined.

We exploit this as follows: Initially all wires are assigned a minimum
rank of~0, since at the beginning we do not know anything about the
values they carry. We then repeat the following construction:
traversing the values $r = k{-}1,\ldots,0$ descendingly, we collect
for each~$r$ all wires with minimum rank~$r$ and apply a splitter
cascade to them (provided there are at least two such wires).
Suppose the wires collected for a minimum rank~$r$ have indices
$[\ell_0,\ldots,\ell_{m(r)-1}]$. After the splitter cascade
we can update the minimum rank of wire~$\ell_i$ to
\hbox{\vbox to0pt{\vss\hbox{$r +2^{b(i)}{-}1$}}}, because
before the splitter cascade there is no known relationship between
wires with the same minimum rank, while the splitter cascade
establishes relationships between them, increasing their ranks by
\hbox{\vbox to0pt{\vss\hbox{$2^{b(i)}{-}1$}}}. The procedure of
traversing the minimum ranks $k{-}1,\ldots,0$ descendingly,
collecting wires with the same minimum rank and applying splitter
cascades to them is repeated until all minimum ranks $0,\ldots,k{-1}$
occur only once.

As an example, Figure~\ref{fig:selnet} shows a $(5,16)$ selection
network constructed is this manner, in which the minimum ranks of the
wires are indicated after certain layers as well as when certain wires
can be discarded (red crosses) and when certain top ranks are determined
(green check marks). Comparators belonging to the same splitter cascade
are shown in the same color.

\begin{table*}[t]
\centering\tabcolsep1.82mm
\begin{tabular}{rrrrrrrrrrrrrrrrrrrrrrrrrrrrrrrrrrrrrrrrrrr}
	\toprule
	$k$   & full &
	\multicolumn{8}{c}{odd-even/pairwise/bitonic selection} &
	\multicolumn{8}{c}{splitter selection} \\ 
	\cmidrule(l){2-2} \cmidrule(l){3-10} \cmidrule(l){11-18}
	\multicolumn{1}{l}{$n$\rule{0pt}{2.3ex}}
	& sort &  1 &  2 &  3 &  4 &  5 &  6 &  7 &  8 
	&  1 &  2 &  3 &  4 &  5 &  6 &  7 &  8 \\
	\midrule
	16 &  10  &  4 &  7 &  9 &  9 & 10 & 10 & 10 & 10
	&  4 &  6 &  7 &  8 & 10 & 11 & 12 & 13 \\
	1024 &  55  & 10 & 19 & 27 & 27 & 34 & 34 & 34 & 34
	& 10 & 14 & 16 & 18 & 22 & 25 & 27 & 29 \\
	10450 & 105  & 14 & 27 & 39 & 39 & 50 & 50 & 50 & 50
	& 14 & 18 & 20 & 23 & 27 & 30 & 32 & 34 \\
	65536 & 136  & 16 & 31 & 45 & 45 & 58 & 58 & 58 & 58
	& 16 & 20 & 22 & 25 & 29 & 32 & 34 & 36 \\ 
	\bottomrule
\end{tabular}
\caption{\label{tab:selnetsizes}Depths of sorting networks and
  selection networks (which are equal for odd-even, pairwise,
  or bitonic networks) compared to selection networks
  constructed with our splitter-based approach. Note that for
  small~$n$ and comparatively large~$k$ an odd-even/pairwise/bitonic
  selection network or even a full sorting network may be preferable
  (e.g.\ $n=16$ and
  $k > 5$), but that for larger~$n$ considerable savings can be
  obtained for small~$k$, even compared to other selection networks.}
\end{table*}

While selection networks resulting from adaptations of sorting networks
(see above) have the advantage that they guarantee that their number
of layers is never greater than that of a full sorting network, our
approach may produce networks with more layers. However, if $k$ is
sufficiently small compared to~$n$ (in particular, if
$k \le \log_2(n)$), our approach can produce selection networks with
considerably fewer layers, as is demonstrated in
Table~\ref{tab:selnetsizes}. Since in the context we consider here
we can expect $k \le \log_2(n)$, splitter-based selection networks
are often superior.

\end{document}